\documentclass{article}

\usepackage[preprint]{neurips_2023}
\usepackage[utf8]{inputenc} 
\usepackage[T1]{fontenc}    
\usepackage{url}            
\usepackage{booktabs}       
\usepackage{amsfonts}       
\usepackage{nicefrac}       
\usepackage{microtype}      
\usepackage{xcolor}         
\usepackage{multirow}
\usepackage[hidelinks]{hyperref}
\usepackage{graphicx}
\usepackage{xspace}
\usepackage{geometry}
\usepackage{changepage}
\usepackage{arydshln}

\usepackage{titlesec}

\titleformat{\part}[display]   
  {\normalfont\large\bfseries} 
  {\partname\ \thepart}        
  {1pt}                       
  {}                           

\newcommand{\llm}{\mbox{\textit{H2O-Danube-1.8B}}\xspace}
\newcommand{\llmchat}{\mbox{\textit{H2O-Danube-1.8B-Chat}}\xspace}
\newcommand{\llmnew}{\mbox{\textit{H2O-Danube2-1.8B}}\xspace}

\title{H2O-Danube-1.8B Technical Report}
\author{
Philipp Singer\thanks{The first two authors contributed equally.}
$\quad$Pascal Pfeiffer\footnotemark[1]
$\quad$Yauhen Babakhin 
$\quad$\textbf{Maximilian Jeblick} \\
$\quad$\textbf{Nischay Dhankhar}
$\quad$\textbf{Gabor Fodor} 
$\quad$\textbf{Sri Satish Ambati}\\
H2O.ai \\
\texttt{\{firstname.lastname, sri\}@h2o.ai}
}
\date{}

\begin{document}

\maketitle

\section{Abstract}

We present \emph{H2O-Danube}, a series of small $1.8B$ language models consisting of \llm, trained on $1T$ tokens, and the incremental improved \llmnew trained on an additional $2T$ tokens. Our models exhibit highly competitive metrics across a multitude of benchmarks and, as of the time of this writing, \llmnew achieves the top ranking on Open LLM Leaderboard for all models below the 2B parameter range.
The models follow core principles of LLama 2 and Mistral, and we leverage and refine various techniques for pre-training large language models. We additionally release chat models trained with supervised fine-tuning followed by direct preference optimization. We make all models openly available under Apache 2.0 license further democratizing LLMs to a wider audience economically. 

\textbf{Danube2 base model:} \url{https://huggingface.co/h2oai/h2o-danube2-1.8b-base} \newline
\textbf{Danube2 chat model:} \url{https://huggingface.co/h2oai/h2o-danube2-1.8b-chat}

\section{Introduction}

Research over the past few years has significantly enhanced language models' capabilities, making them pivotal in tasks like text and code generation, question answering, translation, summarization, and more \cite{ye2023comprehensive}. Most state-of-the-art large language models (LLMs) leverage decoder attention architectures \cite{vaswani2017attention} popularized by the series of GPT models \cite{radford2018gpt1, radford2019gpt2, brown2020gpt3} exemplifying the benefits of pre-training such models on extensive text corpora. 

Scaling laws for LLMs suggest that performance scales by factors such as model and dataset size, as well as computational resources for training \cite{kaplan2020scaling}. This has led to the development of a plethora of models, ranging in size to optimize performance given certain data and compute constraints; notable representatives are: Falcon \cite{penedo2023refinedweb}, Llama 2 \cite{touvron2023llama}, Qwen \cite{bai2023qwen}, Mistral \cite{jiang2023mistral}, or Mixtral \cite{jiang2024mixtral}.

Despite the trend towards larger models, smaller LLMs have taking an important place in today's landscape allowing for efficient inference on consumer hardware and edge devices. While larger models often times excel across various generic tasks \cite{touvron2023llama, bai2023qwen, jiang2023mistral}, fine-tuning smaller models for specific tasks can enable competitive performance with benefits of model size and inference speed \cite{fu2023specializing}, a concept also proven by the success of BERT and its derivatives \cite{devlin2018bert, he2020deberta}. 

In this report, we want to extend previous research in this area \cite{biderman2023pythia, zhang2024tinyllama, zhang2022opt, bai2023qwen, stablelm} and present a series of models based on incremental research and training efforts. We release all models with open weights under Apache 2.0. The first part describes the initial \llm model, as trained on $1T$ tokens, and a separation Section~\ref{sec:danube2} describes \llmnew, a continued modeling effort trained on additional $2T$ tokens. In order to transparently elaborate our incremental insights, the first part is identical to an earlier version of this report\footnote{\url{https://arxiv.org/abs/2401.16818v1}}, while the second part highlights new insights of the second iteration.

Fundamentally, H2O-Danube follows a decoder LLM architecture adopting core principles from Llama 2 \cite{touvron2023llama} and Mistral \cite{jiang2023mistral}.
The models are trained on a combination of, but not limited to, web documents, encyclopedia and public knowledge databases, excluding coding data. \llmnew is trained on a  a more diverse mix of data over multiple data stages. Compared to recent models released in this parameter range \cite{bai2023qwen, zhang2024tinyllama, stablelm}, our models demonstrate to be highly competitive across various benchmarks. As of this writing, \llmnew is the highest ranked open model on the Hugging Face Open LLM Leaderboard\footnote{\url{https://huggingface.co/spaces/HuggingFaceH4/open_llm_leaderboard}} for models below the 2B range. Alongside the base modes, we release chat variants, enhanced with supervised fine-tuning on instruction data and preference data optimization (DPO).

\section{Model architecture}
\label{sec:architecture}

We adjust the Llama 2 architecture \cite{touvron2023llama} for a total of around 1.8b parameters with a hidden size of $2,560$, an intermediate size of $6,912$, and a total of $24$ hidden layers. We use the original Llama 2 tokenizer with a vocabulary size of $32,000$ and train our model up to a context length of $16,384$ (see Section~\ref{sec:training}). In the following, we elaborate more details about the architecture of \llm.

\noindent\textbf{Sliding window.} We adopt the sliding window approach for local attention popularized by Mistral \cite{jiang2023mistral} as implemented in FlashAttention-2 \cite{dao2022flashattention}. For training, we use a fixed sliding window of $4,096$.

\noindent\textbf{Rotary Positional Embedding.} To model dependencies of elements at different positions of a sequence, we use the Rotary Positional Embedding (RoPE) technique as introduced by Su et al. \cite{su2024roformer} and successfully being applied in multiple popular foundation models \cite{touvron2023llama,jiang2023mistral}.

\noindent\textbf{Grouped-query attention.} For reducing the memory bandwidth overhead, we utilize grouped-query attention \cite{ainslie2023gqa}, setting our architecture to use $32$ attention heads and $8$ key-value heads.

\noindent\textbf{Further details.} We rely on root mean square layer normalization (RMSNorm) \cite{zhang2019root} separately for pre- and post-normalization to stabilize training as commonly used in modern LLMs \cite{touvron2023llama}. We do not use bias within linear layers nor tie word embeddings.

\begin{figure}[ht!]
    \centering
    \includegraphics[width=0.48\linewidth]{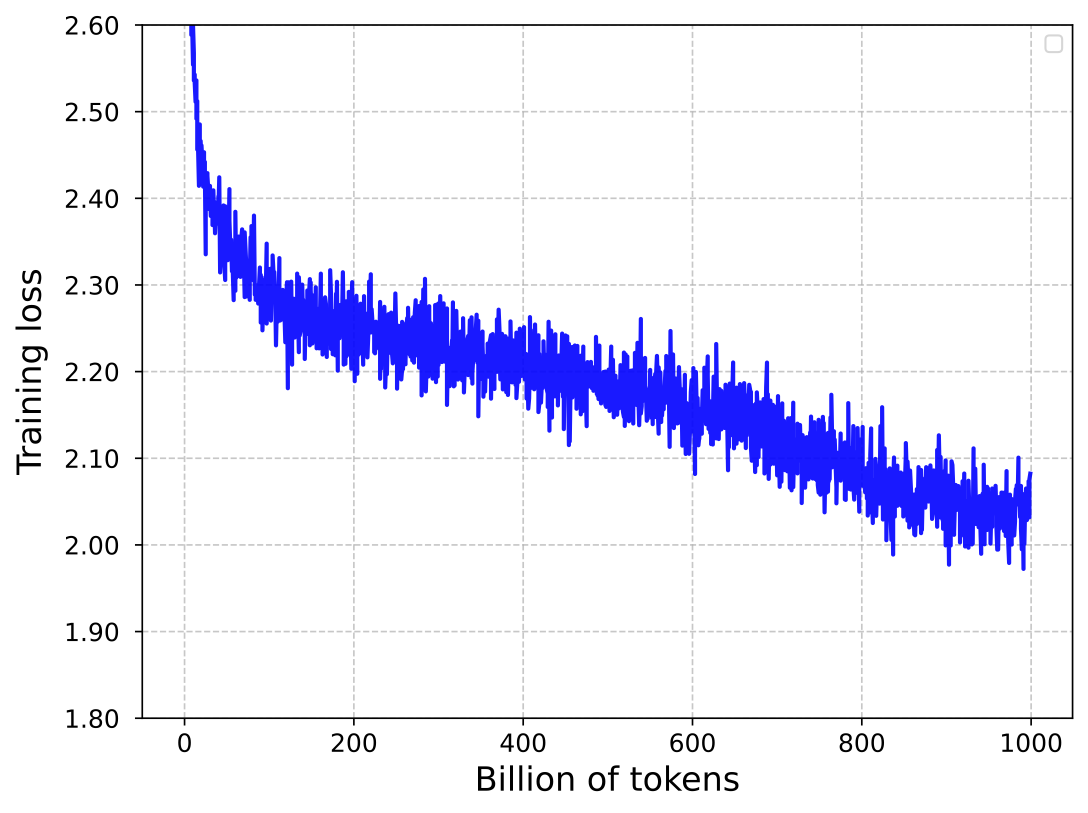}
    \includegraphics[width=0.48\linewidth]{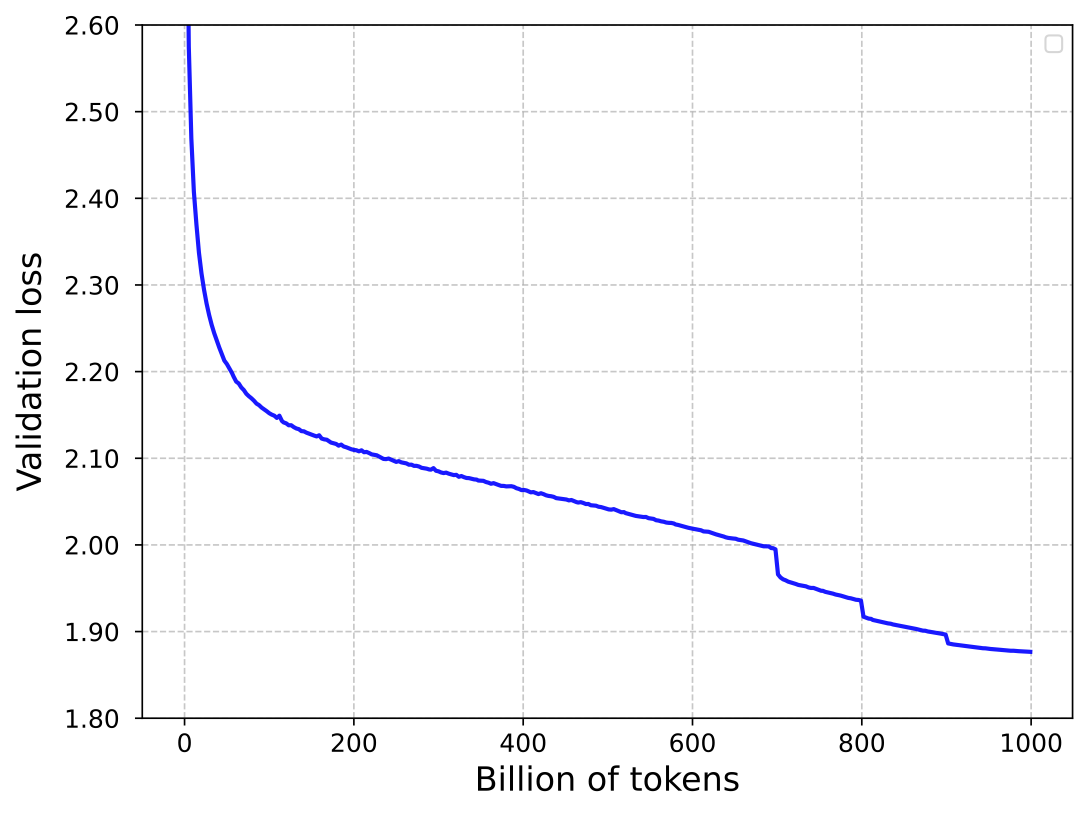}
    \includegraphics[width=0.48\linewidth]{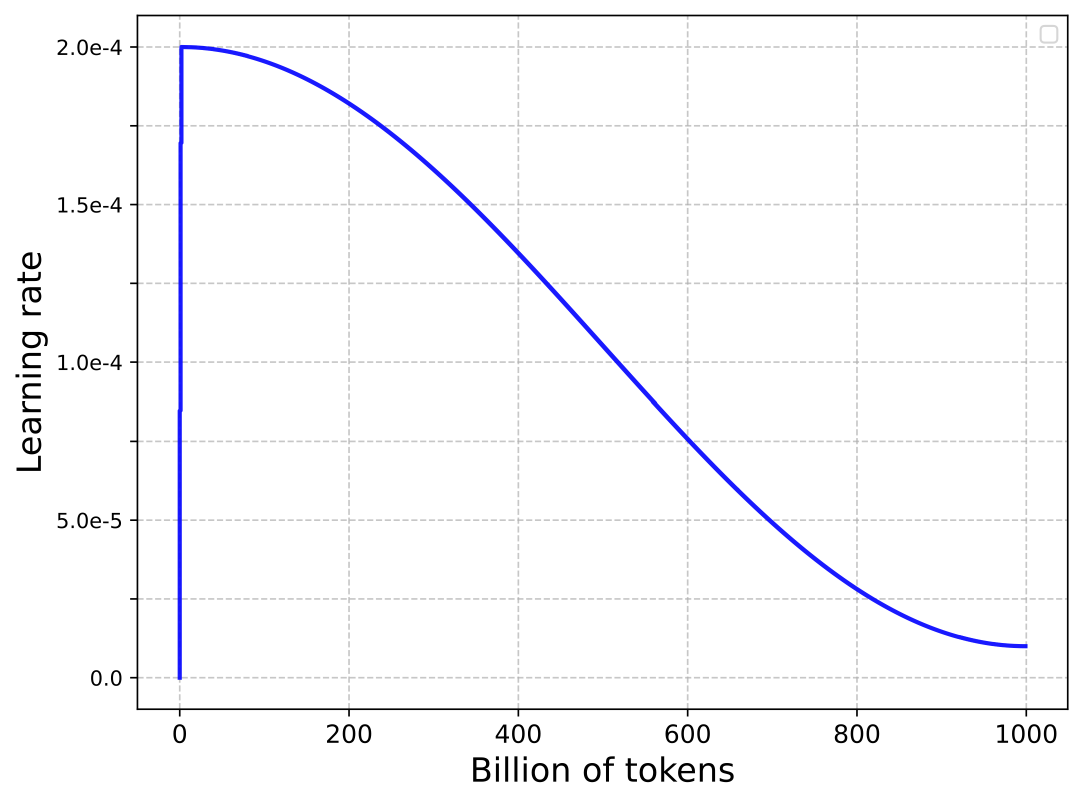}
    \includegraphics[width=0.48\linewidth]{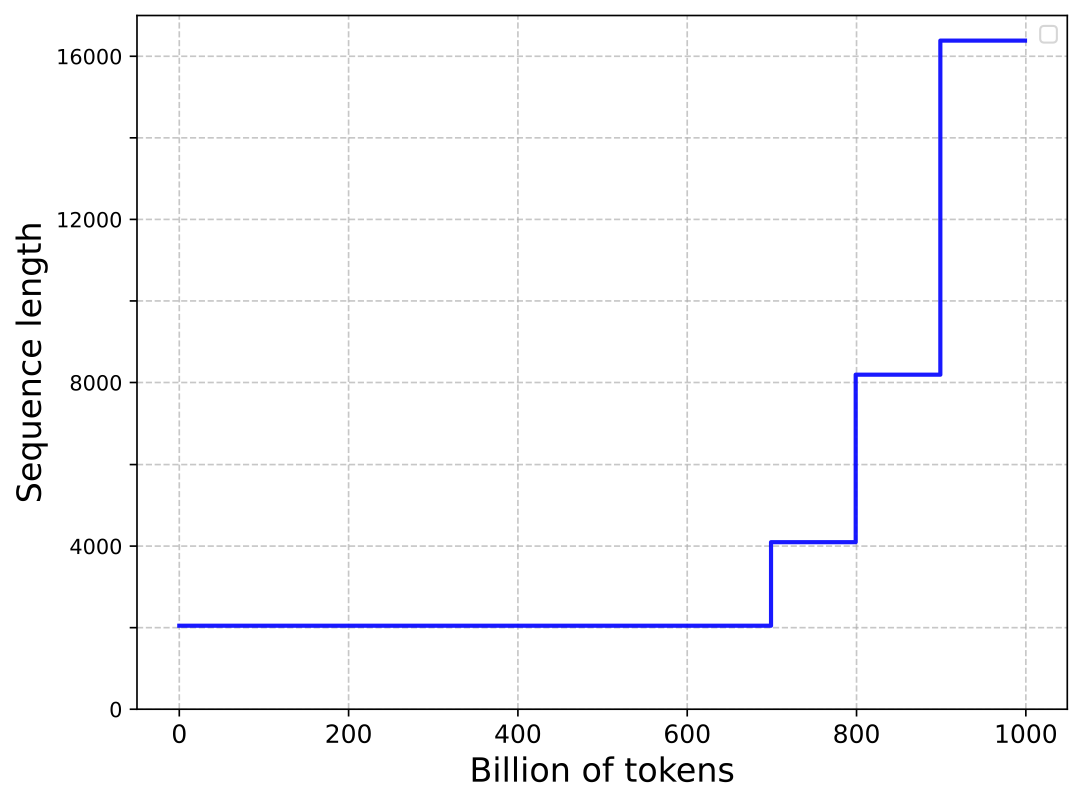}
    \caption{\textbf{Training logs.} Training (top left) and validation (top right) cross-entropy loss, learning rate schedule (bottom left) and sequence length (bottom right). X-axis shows the number of tokens that have been trained up to the step.}
    \label{fig:train_val_loss_old}
\end{figure}

\section{Training}
\label{sec:training}

We train on a single node consisting of 8xH100 GPUs. With Distributed Data Parallel (DDP), each GPU holds a full copy of the model. For finding good settings for our training routine and hyperparameters, we conducted initial experiments on smaller subsets of the data and model sizes up to $500M$ parameters.

Among other findings, these initial experiments showed, that for higher token throughput and compute efficiency, we can iteratively increase the sequence length during the training using a constant sliding window of 4,096 (see Section~\ref{sec:architecture}). Out of the $1T$ tokens in total, we train subsequently on

\begin{itemize}
  \item $700B$ tokens with a sequence length of $2,048$,
  \item $100B$ tokens with a sequence length of $4,096$,
  \item $100B$ tokens with a sequence length of $8,192$,
  \item $100B$ tokens with a sequence length of $16,384$.
\end{itemize}

We employ recent advances in 8-bit floating-point (FP8) calculations on the Hopper architecture to further speed up the training. For this, we cast all linear layers in the Grouped-Query Attention and in the Multi-Layer Perceptron to FP8. The \verb|lm_head| layer remains in bfloat16 precision to ensure training stability. 

We use AdamW optimizer \cite{loshchilov2017decoupled} with $\beta_1=0.9$ and $\beta_2=0.95$ as well as a cosine learning rate scheduler (see Figure~\ref{fig:train_val_loss_old}). We warm up the learning rate for ${\sim}2.36B$ tokens to a maximum of $2e-4$ and then decay it to a minimum of $1e-5$. Our total batch size across GPUs is ${\sim}1.18M$ tokens, weight decay is $1.e-1$ and gradient clipping threshold is set to $1.0$. With these settings, we achieved an average throughput of $292.7k$ tokens/s on the single node during training.

\section{Results}
\label{sec:results}

We evaluate \llm on a wide range of benchmarks and compare it with other existing open-source language models which have a similar number of parameters. Such models include TinyLlama with $1.1B$ parameters \cite{zhang2024tinyllama}, Falcon with $1.3B$ parameters \cite{penedo2023refinedweb}, OPT with $1.3B$ and $2.7B$ parameters \cite{zhang2022opt}, Cerebras-GPT with $1.3B$ and $2.7B$ parameters \cite{dey2023cerebrasgpt}, Pythia-deduped with $1.4B$ and $2.8B$ parameters \cite{biderman2023pythia}, Qwen with $1.8B$ parameters \cite{bai2023qwen}, and most recent Stable LM 2 with $1.6B$ parameters \cite{stablelm}. The majority of these models have Apache 2.0 license, however, Stable LM 2 and Qwen require additional conditions for commercial use and are marked with an asterisk in Table~\ref{tab:commonsense}.

\begin{table}[b!]
\caption{ 
    \textbf{Commonsense reasoning, world knowledge and reading comprehension benchmarks.} \llm exhibits consistently good results across all the benchmarks compared to other models of a similar size. It shows better performance than Qwen on all the benchmarks except for BoolQ, being of the same size but trained on 2.2 times fewer tokens. Stable LM 2 slightly outperforms \llm on the majority of the benchmarks, but was trained on four times the number of tokens. Moreover, neither Qwen nor Stable LM 2 models have the Apache 2.0 license requiring additional conditions for commercial use.
  }
\begin{adjustwidth}{-1in}{-1in}
  \centering
  \small
  \begin{tabular}{lcccccccccc}
    \toprule
  \textbf{Model} & \textbf{Size} & \textbf{Tokens} & \textbf{ARC-e} & \textbf{ARC-c} & \textbf{Bool} & \textbf{HS} & \textbf{OB} & \textbf{PIQA} & \textbf{Triv} & \textbf{Wino} \\
     & & & acc\_n & acc\_n & acc & acc\_n & acc\_n & acc\_n & em & acc \\
    \midrule  
     \multirow{1}{*}{TinyLlama} & 1.1B & 3.00T & 55.35 & 30.12 & 57.80 & 59.18 & 36.00 & 73.18 & 28.78 & 58.88 \\\midrule
    \multirow{1}{*}{Falcon} & 1.3B & 0.35T & 57.32 & 32.00 & 62.84 & 61.52 & 35.20 & 74.65 & 27.27 & 60.77 \\\midrule
    \multirow{2}{*}{OPT} & 1.3B & 0.18T & 50.93 & 29.52 & 57.71 & 53.73 & 33.40 & 72.52 & 15.67 & 59.59 \\
      & 2.7B & 0.18T & 54.34 & 31.31 & 60.31 & 60.61 & 35.20 & 74.81 & 22.38 & 60.85 \\\midrule
    \multirow{2}{*}{Cerebras} & 1.3B & 0.03T & 45.88 & 25.26 & 59.36 & 38.37 & 29.20 & 66.76 & 05.49 & 52.01 \\
      & 2.7B & 0.05T & 52.53 & 27.30 & 59.20 & 48.84 & 32.00 & 70.89 & 12.41 & 55.96 \\\midrule
    \multirow{2}{*}{Pythia} & 1.4B & 0.30T & 56.57 & 29.86 & 56.76 & 54.40 & 33.20 & 72.36 & 18.46 & 56.20 \\
    & 2.8B & 0.30T & 58.96 & 32.68 & 64.19 & 59.45 & 35.60 & 73.78 & 24.39 & 58.17 \\\midrule
    \multirow{1}{*}{Qwen*} & 1.8B & 2.20T & 58.25 & 34.98 & 67.13 & 58.82 & 33.40 & 72.85 & 23.92 & 58.96 \\\midrule
    \multirow{1}{*}{Stable LM 2*} & 1.6B & 4.00T & \textbf{67.63} & \textbf{39.08} & \textbf{75.60} & \textbf{68.78} & \textbf{40.00} & 76.39 & 33.84 & \textbf{63.30} \\\midrule
     \multirow{1}{*}{H2O-Danube} & 1.8B & 1.00T & 62.29 & 35.84 & 65.81 & 68.20 & 37.60 & \textbf{76.93} & \textbf{38.99} & 61.96 \\
    \bottomrule
  \end{tabular}
\end{adjustwidth}
    
  \label{tab:commonsense}
\end{table}

To evaluate the models, we use the Language Model Evaluation Harness framework \cite{eval-harness}. Specifically, we use the version of the framework that is utilized in Open LLM Leaderboard \cite{open-llm-leaderboard}. We report model capabilities across a wide variety of benchmark domains: commonsense reasoning, world knowledge,
reading comprehension and an aggregated Open LLM Leaderboard benchmark.

\noindent{\textbf{Commonsense Reasoning.}} In Table~\ref{tab:commonsense}, we present six standard common sense reasoning benchmarks in 0-shot setting: ARC easy and challenge \cite{clark2018think}, HellaSwag \cite{zellers2019hellaswag}, OpenBookQA \cite{mihaylov2018can}, PIQA \cite{bisk2020piqa}, Winogrande \cite{sakaguchi2021winogrande}. 

\noindent{\textbf{World Knowledge.}} We evaluate 5-shot performance on TriviaQA \cite{joshi2017triviaqa} which represents a closed-book question answering benchmark. Results are presented in Table~\ref{tab:commonsense}.

\noindent{\textbf{Reading Comprehension.}}
We report 0-shot performance on BoolQ \cite{clark2019boolq} in Table~\ref{tab:commonsense}.
    
\noindent{\textbf{Open LLM Leaderboard.}} It evaluates models on six key benchmarks which test a variety of reasoning and general knowledge across a wide range of fields in 0-shot and few-shot settings. It consists of the following benchmarks: ARC challenge (25-shot) \cite{clark2018think}, HellaSwag (10-shot) \cite{zellers2019hellaswag}, MMLU (5-shot) \cite{Hendrycks2020MeasuringMM}, TruthfulQA (0-shot) \cite{lin2022truthfulqa}, Winogrande (5-shot) \cite{sakaguchi2021winogrande}, GSM8k (5-shot) \cite{cobbe2021training}. Results are presented in Table \ref{tab:openllm}

\begin{table}[t!]
\caption{ 
    \textbf{Open LLM Leaderboard.} For each model in the table we report all the individual benchmarks, the average score and the average score without GSM8k benchmark. \llm shows the results similar to Qwen and Stable LM 2 models on the majority of the benchmarks apart from GSM8k and MMLU. It can be explained by the data used for model training, for example, Qwen used gsm8k-ScRel dataset \cite{yuan2023scaling} for the better math reasoning.
  }
\begin{adjustwidth}{-1in}{-1in}
  \centering
  \small
  \begin{tabular}{lccccccccc}
    \toprule
  \textbf{Model} & \textbf{Size} & \textbf{ARC} & \textbf{HS} & \textbf{MMLU} & \textbf{TQA} & \textbf{Wino} & \textbf{GSM} &  \textbf{Average} &  \textbf{Average} \\
     & & 25-shot & 10-shot & 5-shot & 0-shot & 5-shot & 5-shot & & no GSM \\
    \midrule  
     \multirow{1}{*}{TinyLlama} & 1.1B & 33.87 & 60.31 & 26.04 & 37.32 & 59.51 & 01.44 & 36.42 & 43.41 \\\midrule
    \multirow{1}{*}{Falcon} & 1.3B & 35.07 & 63.56 & 25.28 & 35.96 & 62.04 & 00.53 & 37.07 & 44.38 \\\midrule
    \multirow{2}{*}{OPT} & 1.3B & 29.52 & 54.53 & 24.96 & 38.71 & 59.75 & 00.15 & 34.60 & 41.49 \\
      & 2.7B & 33.96 & 61.43 & 25.43 & 37.43 & 61.96 & 00.23 & 36.74 & 44.04 \\\midrule
    \multirow{2}{*}{Cerebras} & 1.3B & 26.28 & 38.54 & 26.59 & 42.70 & 53.43 & 00.23 & 31.30 & 37.51 \\
      & 2.7B & 29.10 & 49.29 & 25.17 & 41.37 & 54.14 & 00.45 & 33.25 & 39.81 \\\midrule
    \multirow{2}{*}{Pythia} & 1.4B & 32.68 & 54.96 & 25.56 & 38.66 & 57.30 & 00.83 & 35.00 & 41.83 \\
    & 2.8B & 36.26 & 60.66 & 26.78 & 35.56 & 60.22 & 00.83 & 36.72 & 43.90 \\\midrule
    \multirow{1}{*}{Qwen*} & 1.8B & 37.71 & 58.87 & \textbf{46.37} & \textbf{39.41} & 61.72 & \textbf{24.41} & 44.75 & 48.82 \\\midrule
    \multirow{1}{*}{Stable LM 2*} & 1.6B & \textbf{43.52} & \textbf{70.45} & 39.08 & 36.81 & \textbf{65.82} & 17.36 & \textbf{45.51} & \textbf{51.14} \\\midrule
     \multirow{1}{*}{H2O-Danube} & 1.8B & 39.68 & 69.75 & 25.97 & 33.63 & 64.17 & 02.05 & 39.21 & 46.64 \\
    \bottomrule
  \end{tabular}
\end{adjustwidth}

  \label{tab:openllm}
\end{table}

For each model in Table \ref{tab:commonsense} we report its number of parameters and the total number of tokens it was trained on. \llm achieves good results across all the commonsense reasoning, world knowledge and reading comprehension benchmarks compared to other models of a similar size. The closest competitors are Qwen and Stable LM 2 models. \llm shows better performance than Qwen on all the benchmarks except for BoolQ. Qwen model has the same 1.8B parameters but was trained on 2.2 times more tokens -- 2.2T. At the same time, \llm is slightly worse than Stable~LM~2 on the majority of the benchmarks, while Stable~LM~2 was trained on four times more tokens -- 2T tokens for 2 epochs. Also, it is important to note that neither Qwen nor Stable LM 2 models have the Apache 2.0 license requiring additional conditions for commercial use.

Similarly, \llm, Qwen and Stable LM 2 are the strongest models on Open LLM Leaderboard (see Table \ref{tab:openllm}) having comparable results on the majority of the benchmarks except for MMLU and GSM8k. A potential explanation for such a behavior might be specifically tailored data that was used for training of Qwen and Stable LM 2 models improving some particular benchmarks, for example, Qwen used gsm8k-ScRel dataset \cite{yuan2023scaling} for better math reasoning.

\section{Chat Fine-Tuning}
\label{sec:chat}

One of the most common use cases for LLMs evolves around instructing and chatting. We thus also provide a chat fine-tuned version \llmchat released under Apache 2.0. We utilize \emph{H2O LLM Studio}\footnote{\url{https://github.com/h2oai/h2o-llmstudio}}, an Apache 2.0 open-source framework and no-code GUI for fine-tuning LLMs.

\subsection{Supervised fine-tuning} 

As first step, we tune the base model using supervised fine-tuning (SFT) on input/output conversational pairs. In detail, we combine the following datasets totalling $157k$ samples: OpenOrca \cite{openorca} following the steps outlined in Orca \cite{mukherjee2023orca}, MetaMathQA \cite{yu2023metamath}, UltraChat200k \cite{ultrachat200k, ding2023enhancing}, and Oasst2 \cite{oasst2}.

We train all layers of the model for a single epoch using a learning rate of $1e-5$, a batch size of $8$, and using cosine learning rate scheduling with a short warmup. We use the full pre-trained context length of $16,384$, mask the prompt loss, and use a custom prompt format. Hyperparameters were optimized iterating over multiple experiments. 

\subsection{DPO}

We follow SFT, by direct preference optimization (DPO) \cite{rafailov2023direct} using a combination of the following datasets: UltraFeedback Binarized \cite{cui2023ultrafeedback}, Orca DPO Pairs \cite{orcadpo} and Distilabel Math Preference DPO \cite{distilabelmathdpo}.
The DPO model is trained using LoRA \cite{hu2021lora} with $r=4$, $alpha=$16 for one epoch using a batch size of $2$ with a learning rate of $1e-5$ using cosine learning rate decay, and $beta=0.2$ for DPO loss.

Afterwards, we run a final DPO fine-tune using Oasst2 \cite{oasst2} dataset building preference pairs from ranks where the chosen answer is the lowest rank, and the rejected answer is the highest one, limited to only English conversations totalling around $5k$ samples. The training run uses similar hyperparameters as the previous one, just a lower learning rate of $3e-6$.

\subsection{Evaluation}

Evaluating chat and instruct fine-tuned LLMs remains a critical challenge and can most reliably be conducted by large scale human assessment. In order to give an initial evaluation of our chat model, we resort to \emph{MT-Bench}, a collection of multi-turn questions across different categories followed by judgement by GPT4 \cite{zheng2023judging}. We keep all categories apart from coding which is out of scope for \llm. Each model is run with $repetition\_penalty=1.1$ and $temperature=0.0$ to reduce randomness and a more fair comparison between models.

We compare our results to popular chat models below $2B$ parameters and highlight them in Table~\ref{tab:mt-bench} showing that \llmchat is exhibiting strong results across categories, particularly for natural language tasks as focused on here. For single turn conversations, \llmchat is the best model for five out of seven categories, and on average on-par with Stablelm 2. For turn 2, we can see that it is comparable to Qwen 2, while Stablelm 2 outperforms other models. 

We make the intermediate sft version\footnote{\url{https://huggingface.co/h2oai/h2o-danube-1.8b-sft}} as well as the final DPO model weights\footnote{\url{https://huggingface.co/h2oai/h2o-danube-1.8b-chat}} available online.
We plan on exploring further improvements for the chat version in the future, working on SFT as well as improved DPO. Particularly, we aim at enhancing preference data with multi turn conversations. We also hope for the open source community to further fine-tune our models for various use cases.

Additionally, we also evaluate chat models on commonsense reasoning, world knowledge, reading comprehension and aggregated Open LLM Leaderboard benchmarks. Similarly as for base models, we report 0-shot benchmark results of the chat versions of \llm, TinyLlama, Qwen and Stable LM 2 in Table \ref{tab:commonsensechat}, and Open LLM Leaderboard results are available in Table \ref{tab:openllmchat}. We show that \llmchat and Stablelm-2-Zephyr perform better than Qwen-Chat and TinyLlama-Chat models on the majority of the benchmarks while being on par between each other. Only exceptions are, again, MMLU and GSM8k benchmarks. As we mentioned in Section~\ref{sec:results}, one of the potential explanations for the worse \llm performance might be a specifically tailored data that was used for training of Qwen and Stable LM 2 base models to optimize those benchmarks.

\begin{table}[hb!]
\caption{ 
    \textbf{Mt-bench chat benchmark.} Both turn 1 and 2 evaluations for mt-bench (ex. coding category) highlight the excellent performance of \llmchat, particularly for single turn conversations showing the highest Mt-bench scores for multiple categories and the average.
  }
\begin{adjustwidth}{-1in}{-1in}
\centering
\small
\begin{tabular}{lcccc}
\toprule
 & TinyLlama-1.1B-Chat & Qwen-1.8B-Chat & Stablelm-2-Zephyr-1.6B & H2O-Danube-1.8B-Chat \\
\midrule
\multicolumn{5}{c}{Turn 1} \\
\midrule
Extraction & 2.50 & 4.70 & \textbf{5.20} & 3.40 \\
Humanities & 8.15 & 6.60 & \textbf{9.20} & 8.90 \\
Math & 1.20 & 2.40 & 3.50 & \textbf{3.80} \\
Reasoning & 3.10 & 3.50 & 3.50 & \textbf{4.30} \\
Roleplay & 5.60 & 6.70 & 6.80 & \textbf{7.05} \\
STEM & 6.60 & 6.50 & 7.35 & \textbf{8.10} \\
Writing & 8.65 & 9.20 & \textbf{9.35} & \textbf{9.35} \\
\\[-2ex]
\hline & \\[-2ex]
Average & 5.11 & 5.66 & \textbf{6.41} & \textbf{6.41} \\
\midrule
\\[-1ex]
\multicolumn{5}{c}{Turn 2} \\
\midrule
Extraction & 1.20 & 2.40 & \textbf{4.80} & 3.20 \\
Humanities & 8.10 & 7.30 & \textbf{9.70} & 8.90 \\
Math & 1.40 & \textbf{1.60} & \textbf{1.60} & 1.50 \\
Reasoning & 2.30 & \textbf{3.90} & 3.20 & 2.70 \\
Roleplay & 5.60 & 6.70 & \textbf{6.95} & 6.90 \\
STEM & 4.60 & 5.80 & \textbf{6.80} & 6.10 \\
Writing & 2.70 & 4.80 & \textbf{7.50} & 3.10 \\
\\[-2ex]
\hline & \\[-2ex]
Average & 3.70 & 4.64 & \textbf{5.79} & 4.63 \\
\bottomrule
\end{tabular}
\end{adjustwidth}
  \vspace{5pt}
  \label{tab:mt-bench}
\end{table}

\vspace{-0.5em}
\begin{table}[hb!]
    \caption{ 
    \textbf{Commonsense reasoning, world knowledge and reading comprehension benchmarks for chat models.} \llmchat outperforms TinyLlama-Chat and Qwen-Chat models, and is on-par with Stablelm-2-Zephyr on all 0-shot benchmarks for commonsense reasoning.  
  }
\begin{adjustwidth}{-1in}{-1in}
  \centering
  \small
  \begin{tabular}{lcccccccccc}
    \toprule
  \textbf{Model} & \textbf{Size} & \textbf{ARC-e} & \textbf{ARC-c} & \textbf{Bool} & \textbf{HS} & \textbf{OB} & \textbf{PIQA} & \textbf{Triv} & \textbf{Wino} \\
     & & acc\_n & acc\_n & acc & acc\_n & acc\_n & acc\_n & em & acc \\
    \midrule  
    \multirow{1}{*}{TinyLlama-1.1B-Chat} & 1.1B & 54.34 & 33.36 & 60.83 & 60.39 & 37.20 & 74.59 & 29.04 & 59.91 \\\midrule
    \multirow{1}{*}{Qwen-1.8B-Chat} & 1.8B & 49.28 & 32.94 & 67.74 & 54.73 & 34.60 & 71.82 & 19.04 & 59.43 \\\midrule
    \multirow{1}{*}{Stablelm-2-Zephyr-1.6B} & 1.6B & 60.35 & \textbf{39.25} & \textbf{80.18} & \textbf{68.85} & \textbf{39.60} & 74.92 & 31.46 & 64.48 \\\midrule
     \multirow{1}{*}{H2O-Danube-1.8B-Chat} & 1.8B & \textbf{67.51} & \textbf{39.25} & 77.89 & 67.60 & 39.20 & \textbf{76.71} & \textbf{36.29} & \textbf{65.35} \\
     
    \bottomrule
  \end{tabular}
\end{adjustwidth}
    \vspace{5pt}

  \label{tab:commonsensechat}
\end{table}

\vspace{-0.5em}
\begin{table}[hb!]
\caption{ 
    \textbf{Open LLM Leaderboard for chat models.} \llmchat outperforms TinyLlama-Chat, and shows similar results to Qwen-Chat and Stablelm-2-Zephyr models apart from GSM8k and MMLU, as also already imminent from results on base models discussed in Table~\ref{tab:openllm}.
  }
\begin{adjustwidth}{-1in}{-1in}
  \centering
  \small
  \begin{tabular}{lccccccccc}
    \toprule
  \textbf{Model} & \textbf{Size} & \textbf{ARC} & \textbf{HS} & \textbf{MMLU} & \textbf{TQA} & \textbf{Wino} & \textbf{GSM} &  \textbf{Average} \\
     & & 25-shot & 10-shot & 5-shot & 0-shot & 5-shot & 5-shot & \\
    \midrule  
    \multirow{1}{*}{TinyLlama-1.1B-Chat} & 1.1B & 36.01 & 61.05 & 25.04 & 37.86 & 60.77 & 02.35 & 37.18 \\\midrule
    \multirow{1}{*}{Qwen-1.8B-Chat} & 1.8B & 36.95 & 54.34 & \textbf{44.55} & 43.70 & 58.88 & 19.26 & 42.94 \\\midrule
    \multirow{1}{*}{Stablelm-2-Zephyr-1.6B} & 1.6B & \textbf{43.69} & \textbf{69.34} & 41.85 & \textbf{45.21} & 64.09 & \textbf{35.18} & \textbf{49.89} \\\midrule
     \multirow{1}{*}{H2O-Danube-1.8B-Chat} & 1.8B & 41.47 & 68.02 & 33.49 & 40.82 & \textbf{64.40} & 15.54 & 43.96 \\
    \bottomrule
  \end{tabular}
\end{adjustwidth}
  \vspace{5pt}

  \label{tab:openllmchat}
\end{table}

\newpage
\section{H2O-Danube2-1.8B}
\label{sec:danube2}

In our effort to grow the ecosystem of permissive open-source foundation models, we publish a new set of models called \llmnew. The base model was initialized from \llm and trained for additional $2T$ tokens. This second iteration of H2O-Danube is the result of extensive experimentation on smaller models, and significantly improves the performance.

\begin{figure}[ht!]
    \centering
    \includegraphics[width=0.48\linewidth]{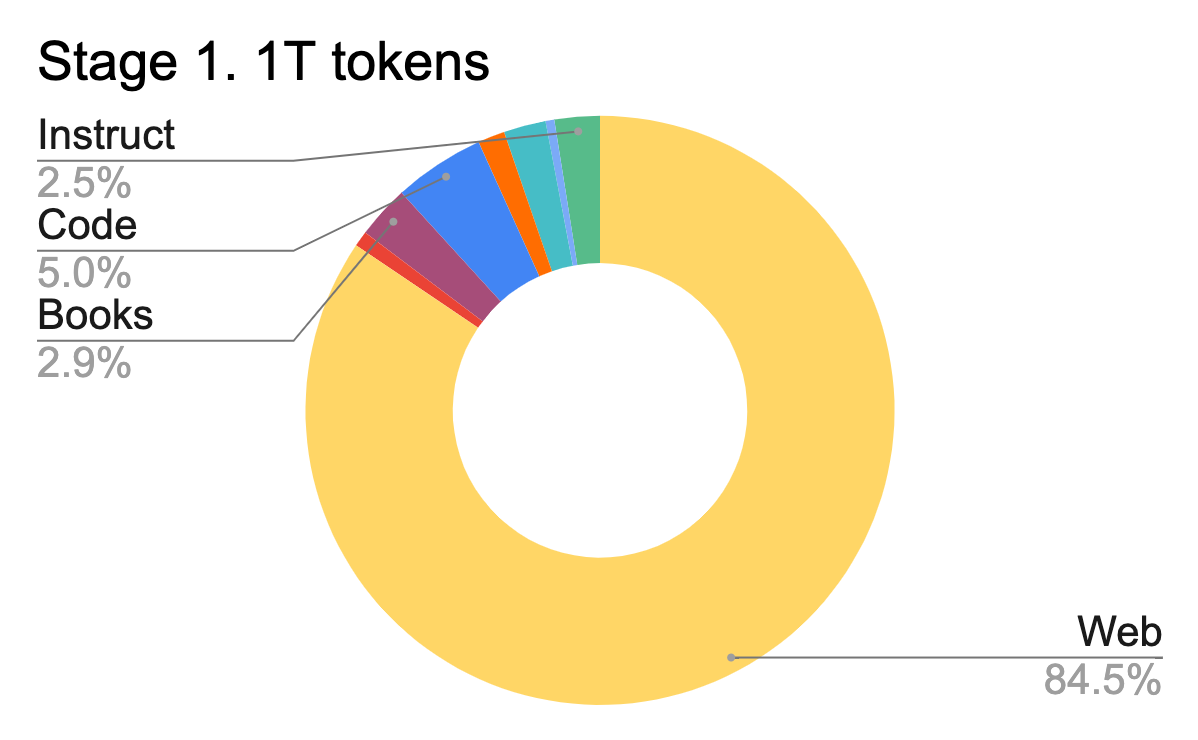}
    \includegraphics[width=0.48\linewidth]{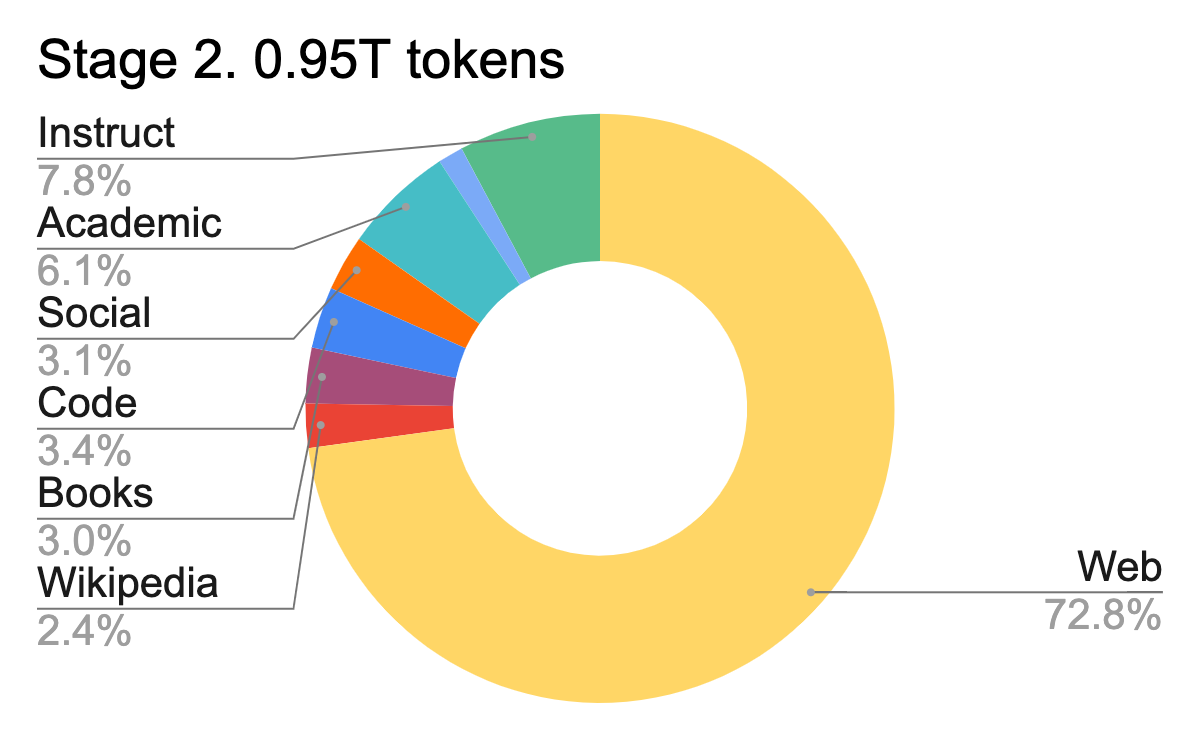}
    \includegraphics[width=0.48\linewidth]{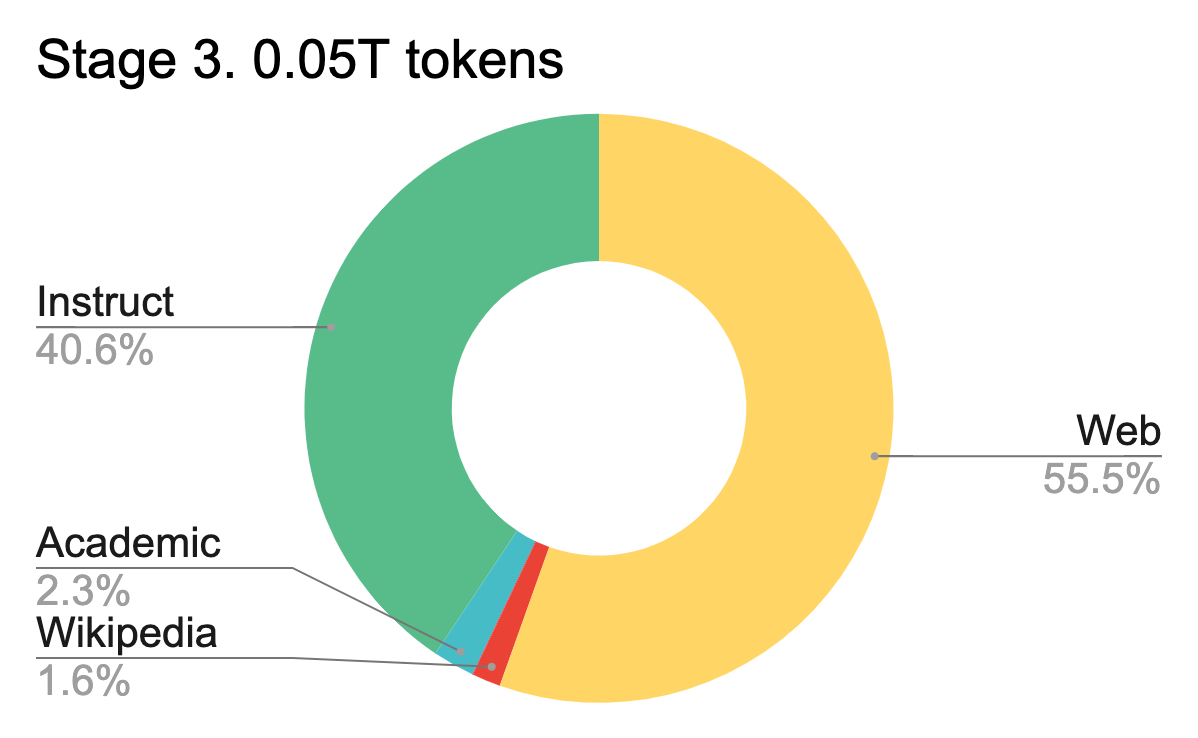}
    \caption{\textbf{Data stages for Danube2.} The model is trained over three different stages with different data mixes. The first data stage consist of 84.5\% of web data which is gradually decreasing to 72.8\% at the second stage, and to 55.5\% at the third stage. The first two stages include the majority of the tokens: 1T and 0.95T tokens respectively, while third stage comprises of 0.05T tokens.}
    \label{fig:data_stages}
\end{figure}

The most significant changes that we have made compared to \llm include:
\begin{itemize}
\item Removal of sliding window attention and change of the maximum context length to $8,192$. By doing so, we effectively improve the long context behavior of the model while keeping memory footprint similar.
\item Change the tokenizer to Mistral which showed superior performance in our experimentation. Instead of fully re-training the embedding and head layers, we re-map the matching tokens and only randomly re-initialize the new tokens.
\item We improve the quality of underlying training data by applying heuristics as well as small models (GBM and BERT) predicting the quality of respective input samples.
\item Training the model in three stages with different data mixes. At each stage, we gradually decrease the percentage of noisy web data in favor of higher quality data. The first data stage consist of 84.5\% of web data which is gradually decreasing to 72.8\% at the second stage, and to 55.5\% at the third stage. Simultaneously, the share of instruct data, Wikipedia, academic texts and other higher quality textual data is increasing. The first two stages include the majority of the tokens: $1T$ and $0.95T$ tokens respectively, while third stage comprises of $0.05T$ tokens. The data distribution across stages is presented in Figure~\ref{fig:data_stages}.
\end{itemize}

Given these adjustments and the continuous training of $2T$ additional tokens, we were able to significantly improve the performance of H2O-Danube.
Since \llm release, there were a couple of new open-weights released in the small models space. For the comparison of base models, we will be using the leading models from Open~LLM~Leaderboard~\cite{open-llm-leaderboard} in the category of $\sim$1.5B parameters (up to 2B parameters); namely, Phi-1.5~\cite{li2023textbooks}, Qwen1.5-1.8B~\cite{bai2023qwen} and StableLM2-1.6B~\cite{stablelm}. We are also comparing to Gemma-2B~\cite{gemma_2024} with 2.5B parameters.
We report OpenLLM Leaderboard results in Table \ref{tab:openllmv1.5}. We can see, that in comparison to the first iteration reported in Table~\ref{tab:openllm}, we can improve on all benchmarks significantly.
As of this writing, \llmnew\footnote{\url{https://huggingface.co/h2oai/h2o-danube2-1.8b-base}} is the highest scoring open model as measured by the average used for the official ranking.

On top of an improved base model, we were also able to develop better chat models following the concepts as described in Section~\ref{sec:chat}. We make the intermediate sft version\footnote{\url{https://huggingface.co/h2oai/h2o-danube2-1.8b-sft}} as well as the final DPO model weights\footnote{\url{https://huggingface.co/h2oai/h2o-danube2-1.8b-chat}} available online. The final \emph{MT-Bench} across all categories and as calculated in the official repository results in a score of $6.23$ for the first turn, $5.34$ for the second turn, and a final average score of $5.79$. 

\begin{table}[hb!]
\caption{ 
    \textbf{Danube2 Open LLM Leaderboard.} For each model in the table we report all the individual benchmarks and the average score. \llmnew achieves state-of-the-art results on this Leaderboard on the average of all benchmarks.
  }
\begin{adjustwidth}{-1in}{-1in}
  \centering
  \small
  \begin{tabular}{lcccccccc}
    \toprule
  \textbf{Model} & \textbf{Size} & \textbf{ARC} & \textbf{HS} & \textbf{MMLU} & \textbf{TQA} & \textbf{Wino} & \textbf{GSM} &  \textbf{Average} \\
     & & 25-shot & 10-shot & 5-shot & 0-shot & 5-shot & 5-shot & \\\midrule
    \multirow{1}{*}{Stable LM 2} & 1.6B & 43.34 & 70.45 & 38.95 & 36.78 & 64.56 & 17.44 & 45.25 \\\midrule
    \multirow{1}{*}{Gemma-2B} & 2.5B & 48.46 & 71.65 & 41.68 & 33.13 & 66.77 & 17.36 & 46.51 \\\midrule
    \multirow{1}{*}{Qwen1.5} & 1.8B & 37.88 & 61.42 & \textbf{46.71} & 39.43 & 60.30 & \textbf{33.59} & 46.55 \\\midrule
    \multirow{1}{*}{Phi-1.5} & 1.3B & \textbf{52.90} & 63.79 & 43.89 & \textbf{40.89} & \textbf{72.22} & 12.43 & 47.69 \\\midrule
     \multirow{1}{*}{H2O-Danube} & 1.8B & 39.42 & 69.58 & 25.94 & 33.86 & 64.48 & 01.44 & 39.12 
      \\
    \multirow{1}{*}{H2O-Danube2} & 1.8B & 43.34 & \textbf{72.95} & 40.20 & 38.01 & 68.03 & 29.80 & \textbf{48.72} \\
    \bottomrule
  \end{tabular}
\end{adjustwidth}

  \label{tab:openllmv1.5}
\end{table}

\section{Conclusions}

We introduce H2O-Danube, a series of new open-source pre-trained foundation model with $1.8B$ parameters including \llm trained on $1T$ tokens and an improved second iteration \llmnew trained on additional $2T$ tokens from diverse sources. The Apache 2.0 license allows for commercial use and for further fine-tuning by the community. We also release a SFT + DPO fine-tuned chat versions, exhibiting state-of-the art results in commonsense reasoning, world knowledge and reading comprehension benchmarks. We show that \llmchat outperforms other models of a similar size on multiple benchmarks.
H2O-Danube is our first contribution to the growing ecosystem of permissive open-source foundation models and we strive to continue publishing high quality foundation models and chat fine-tunes in the near future. Notably, small models can be used on consumer hardware further democratizing LLMs to a wider audience economically.

\small
\bibliographystyle{plain}
\bibliography{references}   

\end{document}